 \documentclass[final,5p,times,twocolumn]{elsarticle}

\usepackage{amssymb}

\usepackage{algorithmic}
\usepackage{algorithm}
\usepackage{bm}
\usepackage{booktabs}
\usepackage{amsmath}
\usepackage{caption}
\usepackage{subcaption}
\usepackage{color}

\usepackage[nodots]{numcompress}

\biboptions{sort&compress}

\journal{Artificial Intelligence in Medicine}

\begin{document}

\begin{frontmatter}



\title{Optimization of anemia treatment in hemodialysis patients via reinforcement learning}

\author[IDAL]{Pablo Escandell-Montero}
\author[FMC]{Milena Chermisi}
\author[IDAL]{Jos\'e M. Mart\'inez-Mart\'inez}
\author[IDAL]{Juan G\'omez-Sanchis}
\author[FMC]{Carlo Barbieri}
\author[IDAL]{Emilio Soria-Olivas}
\author[FMC]{Flavio Mari}
\author[IDAL]{Joan Vila-Franc\'es}
\author[FMC]{Andrea Stopper}
\author[FMC,KREMS]{Emanuele Gatti}
\author[IDAL]{Jos\'e D. Mart\'in-Guerrero}
\address[IDAL]{Intelligent Data Analysis Laboratory, University of Valencia,
Av. de la Universidad s/n, 46100 Burjassot (Valencia), Spain}
\address[FMC]{Healthcare and Business Advanced Modeling, Fresenius Medical Care, Else-Kr\"{o}ner-Strasse 1, 61352 Bad Homburg, Germany}
\address[KREMS]{Centre for Biomedical Technology at Danube, University of Krems, Dr.-Karl-Dorrek-Strasse 30, 3500 Krems, Austria}

\begin{abstract}
\emph{Objective}: Anemia is a frequent comorbidity in hemodialysis patients that can be successfully treated by administering erythropoiesis-stimulating agents (ESAs). ESAs dosing is currently based on clinical protocols that often do not account for the high inter- and intra-individual variability in the patient's response. As a result, the hemoglobin level of some patients oscillates around the target range, which is associated with multiple risks and side-effects. This work proposes a methodology based on reinforcement learning (RL) to optimize ESA therapy.

\noindent \emph{Methods}: RL is a data-driven approach for solving sequential decision-making problems that are formulated as Markov decision processes (MDPs). Computing optimal drug administration strategies for chronic diseases is a sequential decision-making problem in which the goal is to find the best sequence of drug doses. MDPs are particularly suitable for modeling these problems due to their ability to capture the uncertainty associated with the outcome of the treatment and the stochastic nature of the underlying process. The RL algorithm employed in the proposed methodology is fitted Q iteration, which stands out for its ability to make an efficient use of data.

\noindent \emph{Results}: The experiments reported here are based on a computational model that describes the effect of ESAs on the hemoglobin level. The performance of the proposed method is evaluated and compared with the well-known Q-learning algorithm and with a standard protocol. Simulation results show that the performance of Q-learning is substantially lower than FQI and the protocol. When comparing FQI and the protocol, FQI achieves an increment of 27.6\% in the proportion of patients that are within the targeted range of hemoglobin during the period of treatment. In addition, the quantity of drug needed is reduced by 5.13\%, which indicates a more efficient use of ESAs. 

\noindent \emph{Conclusion}: Although prospective validation is required, promising results demonstrate the potential of RL to become an alternative to current protocols.
\end{abstract}


\end{frontmatter}


\section{Introduction}
\label{sec:introduction}

Anemia is a common complication characterized by a reduced concentration of hemoglobin (Hb) that occurs in over 90\% of patients undergoing hemodialysis~\cite{Omara2008}. Hemodialysis is the most common treatment for patients in advanced stages of chronic kidney disease (CKD), particularly in its end state, commonly referred as end-stage renal disease (ESRD). In the last years the prevalence of ESRD has increased substantially, reaching more than 1000 per million population in most of the developed countries~\cite{USRDS2010}. In some countries, such as USA and Japan, the current prevalence is over 2000 per million~\cite{USRDS2010}. ESRD involves a gradual loss of kidney function over time, which produces, among other health problems, a poor production of erythropoietin (EPO). This hormone regulates the red blood cell (RBC) production, a class of cells rich in Hb. Low Hb levels are associated with heart disease, poorer overall quality of life, and increased mortality~\cite{Barany1993,Foley1996}.

Current standard treatment of anemia consists mainly of the administration of erythropoiesis-stimulating agents (ESAs). The response to this kind of drugs is known to have a large inter- and intra-interindividual variability due to differences in background characteristics, disease severity, comorbidities and concurrent medications~\cite{Koch1974,Ifudu1996}. Although there exist protocols to help physicians determine the appropriate dose, achieving stable Hb levels within the target range can be complex and often requires dose titration. Results from several studies suggest that a phenomenon known as Hb cycling is a common occurrence in ESA-treated patients~\cite{Macdougall2005, Collins2005}. Hb cycling is defined as the cyclical, repeated, up and down movement of Hb levels during ESA treatment. The exact causes of Hb cycling are not yet completely understood; however, a number of possible reasons have been proposed. \citet{Fishbane2007} suggested two ESA management practices as major causes. First, the use of rigid dose adjustment protocols that do not account for the high heterogeneity in patient response. Second, narrow Hb target ranges recommended in clinical guidelines~\cite{KDOQI2006,Locatelli2004}, which need frequent dose changes. The effect of an ESA dose change does not reach a steady state until 70-120 days (RBC lifespan). When doses are changed frequently, it is difficult to take into account the long-term effects of each dose, and often they are ignored~\cite{Kalicki2008,Daugirdas2011}. The link between Hb cycling and the development of several diseases~\cite{Macdougall2005} together with the high cost of the treatment (e.g., around \$2.3 billions per year in USA~\cite{USRDS2010}) justifies the need to improve current protocols.

The widespread use of electronic medical records is giving rise to large amounts of data that could be useful to reduce medical errors, improve treatments and minimize side effects and costs~\cite{Patel2009}. This work proposes a methodology based on reinforcement learning (RL) to optimize ESA therapy. RL is a data-driven approach for solving sequential decision-making problems that are formulated as Markov decision processes (MDPs)~\cite{Puterman2005}. Computing optimal drug administration strategies for chronic diseases is a sequential decision-making problem in which the goal is to find the best sequence of drug doses. MDPs are particularly suitable for modeling these problems due to their ability to capture the uncertainty associated with the outcome of the treatment and the stochastic nature of the underlying process~\cite{Hauskrecht2000,Alagoz2010,Bennett2013}. The standard approach to solve MDPs is dynamic programming (DP); however, the practical application of DP is limited because it cannot deal with large-scale problems and requires full knowledge of the MDP model, including the transition probability function. In contrast, RL (also known as approximate dynamic programming (ADP)) uses function approximation to address large-scale problems and the data sampled from the process to implicitly represent the transition function~\cite{Szepesvari2010}. RL can exploit the information contained in medical records to compute policies of ESA administration tailored to the individual characteristics of each patient. In addition, the optimization process is made over sequences of doses instead of isolated doses, which is crucial to include the drug long-term effects. 



The methodology proposed in this work uses the algorithm fitted Q iteration to learn a policy of ESA administration from a set of medical records. The features employed to define the MDP model are extracted in part from the laboratory tests and in part from a clustering procedure of the patient's main attributes. In order to test the methodology, a series of experiments has been conducted using a computational model that simulates the response of the patients. The performance has been assessed against the algorithm Q-learning and a standard protocol of dose adjustment.

The rest of the paper is organized as follows. Next section provides a brief review of related work in this domain. Section~\ref{sec:reinforcementLearning} introduces the necessary background in RL and briefly explains the algorithms employed in the experiments, namely, Q-learning and fitted Q iteration. The latter algorithm makes use of extremely randomized trees, a supervised learning method that is described in Section~\ref{sec:extremelyRandomizedTrees}.
Section~\ref{sec:model} discusses the computational model used in the experiments to simulate patients' response to ESA. Anemia management formulation using the MDP framework is presented in Section~\ref{sec:problemModelling}. Experiments carried out are detailed in Section~\ref{sec:experiments}. Section~\ref{sec:results} shows and discusses the achieved results. Finally, conclusions and proposals for further work are given in Section~\ref{sec:conclusions}.

\section{Literature review}
\label{sec:literatureReview}

The idea of using a data-driven method to optimize ESA administration is not new. Artificial neural networks have been used by several authors during the last decade to individualize ESA doses~\cite{Jacobs2001, Martin2003, Gabutti2006}. In general, those methods used current and previous Hb levels, ESA doses, and other variables  that describe the patient's condition, in order to predict the next Hb level. The goal of those previous works was to select the optimal ESA dose in order to achieve a given Hb level. This approach is suitable only when the optimization horizon is the next time step. On the contrary, the aim of ESA therapy is the long-term Hb stabilization. The same idea has been applied using other machine learning techniques, such as fuzzy logic~\cite{Gaweda2008, Gaweda2003}, support vector machines~\cite{Martin2003b} or Bayesian networks~\cite{Bellazzi1993}. 

Model predictive control (MPC) is a method of process control whose main advantage is that it incorporates a finite time-horizon in the optimization process. \citet{Gaweda2008b} showed that MPC may result in improved anemia management. A major difficulty of MPC is the requirement of an accurate system model. Even if the system model is available, RL has shown to be competitive with MPC~\cite{Ernst2009}.

RL in the context of anemia management was previously studied by~\citet{Gaweda2005} and \citet{Martin2009}. Both agree in the potential of RL to become an alternative to currently used protocols. The algorithm employed in those works was the popular Q-learning~\cite{Watkins1992}. This algorithm has been widely used in some fields as robotics because it requires little computation and can work in real time. However, Q-learning makes an inefficient use of the data, thus, it is not suitable for problems in which acquiring data is costly~\cite{Wiering2012}. Fitted Q-iteration (FQI)~\cite{Ernst2005} is a relatively new RL algorithm that significantly reduces the quantity of data required to learn useful policies. Recently there has been a growing interest in applying FQI to optimize the treatment of several diseases including HIV/AIDS~\cite{Ernst2006}, psychiatric disorders~\cite{Murphy2006}, epilepsy~\cite{Guez2008,Pineau2009}, schizophrenia~\cite{Shortreed2010,Lizotte2012} or smoking addiction~\cite{Chakraborty2008}. To the authors' knowledge, this is the first work that applies FQI to the optimization of anemia treatment.


\section{Reinforcement learning}
\label{sec:reinforcementLearning}
\begin{figure}[]
\begin{center} 
{\includegraphics[width=0.72\columnwidth]{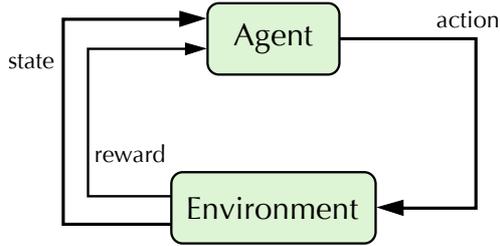}}
\caption{Elements of RL and their flow of interaction.}\label{fig:RLscheme}
\end{center}
\end{figure}

Reinforcement learning (RL) is a general class of algorithms in the field of machine learning for solving decision-making problems where decisions are made in stages~\cite{Sutton1998}. Such problems are present in a wide range of fields, including operations research~\cite{Demircan2011, Bernardo2010}, artificial intelligence~\cite{Peters2008, Gu2007}, automatic control~\cite{Hernandez2012},  or medicine~\cite{Martin2009}. The standard RL setting consists of an agent (or controller) in an environment (or system). Each decision (also called action) produces an immediate reward. The agent learns to perform actions in order to maximize the reward collected over time. The goal is defined by the user through the reward function. Contrary to other approaches, RL does not rely on a mathematical model of the system, but is based on experience (or data). The agent obtains experience interacting with the environment. Fig.~\ref{fig:RLscheme} 
represents the main RL elements and how they interact.
At each stage or discrete time-point $k$, the agent receives the environment's state, and on that basis selects an action. As a consequence of its action, in the next time step, the agent receives a numerical reward and the environment evolves to a new state. The agent selects actions depending on the environment state using a policy that assigns an action to every state. Typically, 
the agent modifies the policy as a result of the interactions with the environment.

The elements of the RL problem can be formalized using the Markov decision processes (MDPs) framework. Next, in Section~\ref{subsec:MDP}, MDPs are used to formally introduce the basic components of RL. Then, the algorithms used in this work, Q-learning and fitted Q iteration, are briefly described in Section~\ref{subsec:solvingMDPs}.

\subsection{Markov decision processes}
\label{subsec:MDP}
Markov decision processes are a general mathematical framework for modeling sequential decision-making problems. An MDP is defined by the following elements:
\begin{itemize}
\item A set of \textbf{states}, $S$: At each discrete time-point $k$, the environment occupies a state $s_k \in S$. The state usually is composed by a vector whose components describe the current situation of the environment. As time passes, the vector values evolve in part as consequence of the actions applied by the agent and in part stochastically. The state can simply be a variable observed directly from the environment, or a more complex structure such as a set of variables highly processed which combines information about the current and the past situations of the environment\footnote{The only requirement is that the state should contain enough information to fulfill the Markov property (see Section~\ref{sec:stateRepresentation} for more details).}.

\item A set of \textbf{actions}, $A$: The agent applies an action $a_k \in A$ in each state. The state in the next instant, $s_{k+1}$, is influenced by the current action. The actions are the mechanism employed by the agent to control or guide the evolution of the environment.
\item A \textbf{transition probability function}, $P: S\times A\times S \rightarrow [0,1)$: After action $a_k$ is taken in the current state, the transition function gives the probability of the next state, i.e., this function describes how the state evolves.
\item A \textbf{reward function}, $\rho: S\times A\times S \rightarrow \mathbb{R}$: Each transition between states generates a reward $r_{k+1}=\rho(s_k,a_k,s_{k+1})$ that evaluates the immediate effect of the transition, but it does not provide information about its long-term effects. The reward function is defined by the user and implicitly codifies the goal of the agent. Notice that the reward function does not describe how to achieve the goal, but the agent must learn how to act from experience. 
\end{itemize}

Suppose a patient suffering from a certain disease that requires long term treatment with a particular drug. Usually, the aim is to administrate a suitable sequence of doses in order to control a variable (or several variables) related to the severity of the disease. For example, in anemic patients, Hb level is used to measure the degree of anemia. The MDP framework can be applied to this problem modeling the patient as the environment. In this case, the state should contain all the information relevant to choose a proper treatment. In addition to the current level of Hb, the state may include other factors that can influence the effect of the treatment such as the physical characteristics of the patients or their nutritional condition. The set of actions are the possible treatments (or drug doses) that can be administered to the patient. After each treatment, the patient status evolves to a new state. The new state will be in part a consequence of the treatment, and in part a consequence of other aspects that cannot be controlled by the agent, like for example the presence of inflammation or blood losses. If the objective of the treatment is to maintain the Hb level within a range, the reward function can be defined to provide a positive reward when the Hb level is between the limits of the target range and a negative reward in otherwise.

The agent selects actions according to its policy $\pi: S \rightarrow A$. The policy is a function that maps states to actions, i.e., for each possible state of the environment, the policy indicates the action that should be performed:
\begin{equation}
\pi(s) = a
\end{equation}
The objective of the agent is to learn a policy that maximizes the sum of rewards received over time, a quantity known as return. Such a maximizing policy, denoted by $\pi^*$, is said to be optimal. The return usually is computed using the infinite-discounted horizon. In such a case, the return for an initial state $s_0$ and under the policy $\pi$ is~\cite{Bertsekas2007}:
\begin{equation}
\label{eq:return}
R^{\pi}(s_0) = \lim_{K \rightarrow \infty }E_{s_{k+1}|s_k,\pi(s_k)}\left\{ \sum_{k=0}^{K} \gamma^k \rho(s_k,\pi(s_k),s_{k+1})\right\} 
\end{equation}
where $\gamma \in [0,1)$ is the discount factor. This parameter can be intuitively interpreted as a way to balance the immediate reward and future rewards. Future rewards are more relevant for the calculation of the return as $\gamma$ approaches 1.

To find an optimal policy, the agent must explore the environment: the probability of attempting new actions (different from those dictated by the policy) must always be non-zero. Otherwise, some areas of the state-action space may never be visited and the learning process can become stuck in a local optimum. The tradeoff between greedy action choices and exploration is necessary for the performance of any RL algorithm~\cite{Busoniu2010}. There exist several strategies to include exploration in the agent's behaviour, such as $\epsilon$-greedy exploration or Boltzmann exploration~\cite{Sutton1998}.

\subsection{Solving MDPs}
\label{subsec:solvingMDPs}
Solving an MDP means to find an optimal policy. There are several methods for solving MDPs, which can be grouped into two classes: dynamic programming (DP) and reinforcement learning (RL). DP methods require knowledge of the full MDP model. Since the transition probability function $P$ rarely is available, this class of methods can only be employed in a limited number of practical problems. On the other hand, RL methods are completely based on experience, which makes them useful when the full MDP model is unknown or difficult to estimate. 

Most RL algorithms use Q-functions (also called utility functions) $Q^\pi: S \times A \rightarrow \mathbb{R}$ to find an optimal policy. Given a policy, the Q-function for a particular pair $(s,a)$ is defined as the expected return that is encountered starting from $s$, taking action $a$ and thereafter following policy $\pi$~\cite{Busoniu2010}: 
\begin{equation}
\label{eq:Qfunction}
Q^{\pi}(s,a) = E_{s^\prime|s,a} \left\{ \rho(s,a,s^\prime) + \gamma R^\pi(s^\prime) \right\}
\end{equation}
where $s^\prime$ is the state reached after taking action $a$ in the state $s$. The Q-function measures the utility (in terms of the expected return) of perform each action in each state.

The optimal Q-function is defined as the best Q-function that can be obtained by any policy:
\begin{equation}
Q^*(s,a) = \max_\pi Q^\pi (s,a)
\end{equation}
From the optimal Q-function, an optimal policy can be easily derived choosing in each state the action that maximizes $Q^*$:
\begin{equation}
\label{eq:policy}
\pi^*(s)= \arg \max_a Q^*(s,a)
\end{equation}
where $\arg \max_x f(x)$ stands for the argument $x$ that attains the maximum value of the function $f(\cdot)$. In general, for a given Q-function, a policy that maximizes $Q$ in this way is said to be greedy in $Q$. Therefore, solving an MDP (i.e., finding an optimal policy) can be done by first finding $Q^*$, and then using Eq.~(\ref{eq:policy}) to compute a greedy policy in $Q^*$. 

When MDPs have a small enough number of states and actions, Q-functions can be exactly stored in tables with one entry per state-action pair. Unfortunately, many practical problems contain a very large or infinite (for example when the state space is continuous) number of states; in such a case, Q-functions must be represented approximately by two reasons. First, suppose that the MDP contains $N$ states, if $N$ is very large, a table with $N$ entries would be intractable due to computational and memory limitations. On the contrary, in a typical approximate representation is only necessary to store a vector of $M$ parameters, being $M \ll N$. Second, when the state space is large or continuous, the agent will probably never be exactly on the same state more than once. Therefore, the experience acquired from a set of states should be generalized to other states that have not been seen before. In principle, any function approximation (or regression) method can be used to represent Q-functions. However, in practice, some RL algorithms impose restrictions about the structure of the approximator.

The remaining of this section describes two RL algorithms for solving MDPs. On the one hand, the well-known Q-learning algorithm~\cite{Watkins1992}. Q-learning is probably the most popular RL algorithm, it has been used in many applications, including the treatment of anemia in hemodialysis patients~\citep{Gaweda2005, Martin2009}. On the other hand, fitted Q iteration~\cite{Ernst2005}, a more recent algorithm that forms part of the methodology proposed in this paper. Both are offline algorithms, which means that they do not require interacting with the environment during the learning phase, but instead can learn a solution using data collected in advance. The data, or experience, usually is stored as a set of transitions of the form $(s,a,r,s')$ sampled from the process. This data set should be representative of the state-action space, i.e., they should contain a certain degree of exploration. Given that the agent cannot interact with the environment when RL algorithms are applied offline, in such a case it is not necessary to include a exploration strategy.

\subsubsection{Q-learning}
 \label{subsubsec:Qlearning}
Consider a Q-function approximator denoted by $F$ and parameterized by a $d$-dimensional vector $\bm{\theta}$. Every possible vector of parameters $\bm{\theta}$ provides an approximated representation of a corresponding Q-function:
\begin{equation}
\hat{Q}(s,a) = F_{\bm{\theta}}(s,a) 
\end{equation}
where the symbol $\hat{\cdot}$ denotes approximation. In general, the approximator $F$ can be nonlinear in the parameters. However, in some algorithms (e.g., Q-learning) linear approximators are often preferred because they provide better convergence and stability properties~\citep{Boyan1995, Sutton1996, Tsitsiklis1997}. A linearly parameterized approximation of the Q-function is expressed as:
\begin{equation}
F_{\theta}(s,a)  = \sum_{l=1}^d \phi_l(s,a)\theta_l = \bm{\phi}^\top(s,a)\bm{\theta}
\end{equation}
where $\bm{\phi}(s,a) = [\phi_1(s,a),\ldots,\phi_d(s,a)]$ is the vector of basis functions (also called features~\citep{Busoniu2010}) that are combined using the $d$-dimensional vector of parameters $\bm{\theta}$. A common approach to define the basis functions consists in using a regular grid of Gaussian radial basis functions (RBFs) spanned over the state-action space~\citep{Sutton1998, Lagoudakis2003}. In such a case, for some state-action pair $x=(s,a)$, the vector of basis functions is:
\begin{equation}
\left[\exp\left(-\frac{||x-c_1||^2}{2\sigma_1^2}\right), \exp\left(-\frac{||x-c_2||^2}{2\sigma_2^2}\right), \ldots, \exp\left(-\frac{||x-c_d||^2}{2\sigma_d^2}\right) \right]^\top
\end{equation}
where each RBF is defined by its position or center $c$ and variance $\sigma^2$.

The Q-learning algorithm starts with an arbitrary approximation of the optimal Q-function, i.e., an arbitrary vector of parameters $\bm{\theta}_0$. Then, it uses the data from each transition $(s_k,a_k,r_{k+1},s'_{k+1})$ to update the parameters using the following rule:
\begin{multline}
\label{eq:aproxQlearning}
\bm{\theta}_{j+1} = \bm{\theta}_j + \alpha \left[ r_{k+1} + \gamma \max_{a'} \bm{\phi}^\top (s_{k+1},a')\bm{\theta}_j \right. \\  \left. -  \bm{\phi}^\top (s_k,a_k)\bm{\theta}_j \right] \bm{\phi} (s_k,a_k) 
\end{multline}
where the index $j = 1, \ldots, p$ corresponds with the number of transitions in the data set and $\alpha$ is the learning rate. This learning rule updates the estimation of the optimal Q-function incrementally. Moreover, each update requires little computation and memory resources, which makes possible to apply Q-learning in real-time. On the other hand, the algorithm presents two possible drawbacks: (i) it generally requires many transitions to obtain useful policies, and (ii) the function approximator should be parametric and typically linearly parameterized~\cite{Busoniu2010}.

 \subsubsection{Fitted Q iteration}
 \label{subsubsec:FQI}
Fitted Q iteration (FQI) is a batch RL algorithm whose main feature lies in the way that it handles the experience~\cite{Ernst2005}. Unlike incremental algorithms, FQI uses the complete set of transitions each time that updates the estimation of the optimal Q-function. Although this process involves more computation, it allows to extract more information from the stored experience. Consequently, FQI is more data-efficient than other RL algorithms. This feature makes FQI a very suitable algorithm in many application domains. For example, in the problem tackled here, patients are modeled as the environment and the agent has to estimate optimal drug doses. Acquiring experience in this context entails administering a dose, waiting until it takes effect and measuring the variables that define the new patient's condition. This process is expensive, both in time and money. Thus, reducing the quantity of data required by the algorithm can be crucial.

Given a fixed set $\mathcal{D}=\{(s^j_k,a^j_k,r^j_{k+1},s^j_{k+1}), j=1, \ldots, p\}$ of $p$ transitions and an arbitrary initial Q-function $Q_0$ (e.g., equal to zero everywhere on $S \times A$), FQI starts by initializing an approximation $\hat{Q}_0$ of the Q-function $Q_0$, with $\hat{Q}_0(s,a)=0$ for all $(s,a) \in S \times A$. It then iterates over the following three steps~\cite{Ernst2005}:
\begin{enumerate}
\item $n \leftarrow n+1$
\item Build the training set $\mathcal{TS}=\{(input^j,target^j),j=1,\ldots, p\}$ based on the function $\hat{Q}_{n-1}$ and on the full set of transitions $\mathcal{D}$:
\begin{align}
\label{}
input^j & =(s_k^j,a_k^j) \\
target^j & =r_{k+1}^j+\gamma \max_{a^\prime} \hat{Q}_{n-1}(s^j_{k+1},a^\prime)
\end{align}
\item Use supervised learning to induce from $\mathcal{TS}$ the function $\hat{Q}_n(s,a)$
\begin{align}
\label{}
\hat{Q}_n(s,a) & =r_{k+1}+\gamma \max_{a^\prime} \hat{Q}_{n-1}(s_{k+1},a^\prime)
\end{align}
\end{enumerate}
The iterative process can be stopped simply by establishing a maximum number of iterations. Another possibility is to fix a threshold value $\xi > 0$ and stop the loop when the distance between two consecutive estimations of the optimal Q-function drops below the threshold, i.e., $|\hat{Q}_n - \hat{Q}_{n-1}| < \xi$~\cite{Ernst2005}.

FQI, similarly to other RL algorithms, requires a function approximator to represent large or continuous Q-functions. However, on the contrary to Q-learning, it does not impose any constraint on the kind of approximator. In fact, at each iteration $n$ it is possible to change the approximator in order to adapt the resolution (or complexity) of the model so as to reach the best bias/variance tradeoff~\cite{Ernst2005}. Although FQI has been successfully combined with many approximation methods (e.g., neural networks~\cite{Riedmiller2005} or linear regression~\cite{Lizotte2012}), a technique known as extremely randomized trees~\cite{Geurts2005} has shown better performance than other approaches~\cite{Guez2008,Ernst2005}. Therefore, this tree-based method, whose details are introduced in the next section, was used in the experiments.

Despite FQI is more data-efficient than other RL algorithms, the number of transitions required to learn optimal policies grows quickly with the state space dimensionality due to the ``curse of dimensionality''~\cite{Bellman1957}. This feature is common to all RL algorithms, including also Q-learning. Thus, reducing the number of state variables as much as possible is an important issue. To this end, similarities among patients were exploited by $k$-means clustering analysis before applying any learning algorithm (see Section~\ref{sec:problemModelling})~\cite{Alpaydin2004}.

\section{Extremely randomized trees}
\label{sec:extremelyRandomizedTrees}
Extremely randomized trees~\cite{Geurts2005} are a tree-based ensemble method for supervised classification and regression problems. It can be considered as an improved version of the popular tree ensemble method Tree Bagging~\cite{Breiman1996}. The algorithm developed to compute this kind of ensembles is called Extra-Trees and, like Tree Bagging, it works by building several ($M$) trees. A key difference between both approaches lies in the sample used to compute the trees. While Tree Bagging uses a bootstrap sample, Extra-Trees uses the complete data set to built each tree.


Similarly to standard regression trees, each tree is composed of decision nodes, where each node contains a split (or test) of an attribute. The value where such attribute is split is known as cut-point. In order to define a decision node, Extra-Trees generates $K$ splits by choosing $K$ attributes at random and for each attribute a cut-point at random. Then, it calculates a score (based on the explained variance) for each of the $K$ candidate splits and selects the split that obtained the maximum score. This process is repeated until the number of elements in the node is less than the parameter $l_{min}$~\cite{Geurts2005}. The algorithm has three parameters that need to be specified: the number $M$ of trees to build the ensemble, the number $K$ of candidate tests at each node and the minimal leaf size $l_{min}$. 
\section{Erythropoiesis model under darbepoetin alfa treatment}
\label{sec:model}

The ability of RL to compute treatment policies was assessed through simulations. Experiments were based on a computational model that describes the effect of darbepoetin alfa on the Hb level. This section presents the main characteristics of the model in order to provide insight into the tackled clinical problem. \ref{sec:appendix} gives a more detailed description of the model.

Several theoretical pharmacodynamic models describing the hematological response to different kinds of ESAs have been developed during the last decades~\cite{Uehlinger1992, Krzyzanski1999, Krzyzanski2005, Perez-Ruixo2005, Krzyzanski2007, Woo2007}. The model introduced in this section is focused in patients undergoing hemodialysis who are treated with intravenous darbepoetin alfa, a second generation ESA drug. The hematopoietic cell populations are natural examples of biological systems governed by lifespan-based processes of cell proliferation, differentiation, maturation, and senescence. The Hb level is proportional to the number of erythrocytes, which are produced primarily from stem cells in bone marrow. During the process of maturation, stem cells undergo a series of differentiations. When they reach the stage of reticulocyte (immature erythrocytes), they begin to circulate in the blood and, after 1-2 days, these ultimately become mature RBCs (erythrocytes). In patients with CKD, erythrocyte lifespan is approximately $70$-$90$ days~\cite{Uehlinger1992}. The process of erythrocytes production (erythropoiesis) is regulated by the hormone EPO, which is produced in the kidneys. Darbepoetin alfa is a synthetic form of EPO that stimulates erythropoiesis by the same mechanism.

Hb concentration dynamics following the administration of darbepoetin can be described through a multi-compartment model~\cite{Jacquez1985}.
The different cell types involved in erythropoiesis are grouped into population classes (or compartments) according to their characteristic properties with respect to interaction with EPO. The number of cells in all compartments depends on the plasma concentration of EPO, which consists of the sum of the naturally produced EPO, $Ep$, and the exogenous administered EPO (darbepoetin alfa), $E$:
\begin{equation}\label{epotot}
E_{tot}= E+Ep.
\end{equation}
The rate of endogenous EPO production in the kidney is assumed to be constant.

In the case of an intravenous administration, the total amount of darbepoetin alfa is injected into a vein within a very short time interval so that, without loss of generality, it can be assumed that the amount of the exogenous hormone in plasma at $t_0$, the time when the administration takes place, is exactly equal to the amount of exogenous hormone administered. The result is a sudden rise of the hormone plasma concentration followed by an exponential decay described by a first order ordinary differential equation with a constant elimination rate
\begin{equation}\label{epoexo}
E^{\prime}(t)= \frac{24}{25}\log(2)E(t), \quad E(t_0)=\frac{D_0}{V_d},
\end{equation}
where $E(t)$ stands for the concentration of exogenous hormone at time $t$, $D_0$ the drug dose at $t_0$, and $V_d$ is the volume of distribution at steady state, which is set to $V_d=52.4$~\cite{Allon2002}.

The total Hb level at time $t$ is proportional to the concentration of erythrocytes $R(t)$:
\begin{equation}\label{hgmodsol}
Hb(t) = \text{MCH} \cdot R(t),
\end{equation}
where $R(t)$ is the plasma concentration of RBCs and MCH denotes the mean corpuscular Hb concentration~\cite{Goldman2011}, chosen as
\begin{equation}
\label{eq:MCH}
\text{MCH} = \left\{
\begin{array}{lr}
	2.7  \text{ ${\rm g}\cdot\big(10^{11}{\rm cells}\big)^{-1}$}&\text{for men.}\\
	2.4  \text{ ${\rm g}\cdot\big(10^{11}{\rm cells}\big)^{-1}$} &\text{for women.}\\
     \end{array}
   \right.
\end{equation}
The complete system of delay differential equations (DDE) that describes the cell dynamics in each compartment is included in~\ref{sec:appendix}. In addition to endogenous EPO ($Ep$) and the mean corpuscolar hemoglobin concentration (MCH), the individual response of each patient to darbepoetin alfa is defined by two other parameters: $Cp$, a constant that determines the flow of cells in the first compartment of the model (denoted by $P$), and $Cr$, which plays a similar role in the last compartment (denoted by $R$) (see~\ref{sec:appendix} for more details). These parameters can be adjusted using clinical and laboratory data of patients. M$\text{\sc{atlab}}$ 7.14 (R2012a) was employed to adjust the parameters and perform simulations. The M$\text{\sc{atlab}}$ DDE solver $dde23$ was used to compute the solutions of the system of differential equations.

This computational model, as any other model, it is a simplification of the real problem. It makes two basic assumptions: first, each patient maintains a stable level of inflammation, and second, the availability of iron for erythropoiesis is constant. Generally, these assumptions are not met by all patients during all treatment. Nevertheless, the model is able to capture the heterogeneity in the response to darbepoetin alfa and its long-term effects, two factors suggested as principal causes of Hb cycling. Therefore, the model is useful to evaluate the performance of RL in preventing Hb cycling.

\section{Problem formulation using the MDP framework}
\label{sec:problemModelling}

\subsection{Anemia management as a sequential decision-making problem}

The symptoms of anemia and the response to the treatment often vary depending on several factors, such as the physical characteristics of the patient (weight, age, sex, etc.)~\cite{Hsu2002}, degree of kidney disease~\cite{Stenvinkel2002} or comorbidities~\cite{Locatelli2006}. During the treatment period, the clinical evolution of each patient is typically monitored by monthly reviews. Each review consists of a laboratory test, which measures several variables to assess the patient's condition, and the drug prescription until the next review. Therefore, each patient generates a sequence of observations and treatments of the form~\cite{Lizotte2012}
\begin{equation}
\label{originalSequence}
o^j_1,t^j_1,o^j_2,t^j_2,\cdots,o^j_T,a^j_T,o^j_{T+1}
\end{equation}
where $o^j_k$ stands for the condition of the $j$-th patient at month $k$, $t^j_k$ stands for the dosage $t$ for the same patient and month, and $T$ is the duration of the whole treatment.

A sequential decision-making problem can be easily derived from this set of sequences, where actions (or decisions) correspond to the doses prescribed to each patient in the different months (or stages). Before applying any RL method, it is necessary to define a function $s(o_1,a_1,\cdots,o_k)$ that assigns a state to the patient's current history and a scalar reward function $\rho(s_k,a_k,s_{k+1})$ that evaluates the transition from $s_k$ to $s_{k+1}$ after taking the action $a_k$. The result of applying both functions to~(\ref{originalSequence}) is a sequence of states, actions and rewards~\cite{Lizotte2012}:
\begin{equation}
\label{RLsequence}
s^j_1,a^j_1,r^j_1,s^j_2,a^j_2,r^j_2,\cdots,s^j_T,a^j_T,r^j_T
\end{equation}
These sequences can be rewritten as a set of transitions $(s,a,r,s^{\prime})$. Then, FQI can be applied to compute a policy that selects optimal drug doses (in view of the data available) depending on the patient's condition.

The rest of this section details the definition of states, actions and reward function in the context of anemia treatment.

\subsection{State and action spaces}
\label{sec:stateRepresentation}
The MDP framework assumes that the state transition dynamics and the reward distribution has the Markov property, that is, given the current state $s_k$ and action $a_k$, the next state $s_{k+1}$ and reward $r_k$ are conditionally independent of all previous states, actions and rewards~\cite{Sutton1998}. An obvious way of defining a state representation that holds the Markov property is to include all past history in $s_k$. In practice, this is not feasible due to two reasons: (i) as stated in Section~\ref{subsubsec:FQI}, the amount of experience needed to learn useful policies grows exponentially with the number of state dimensions due to the curse of dimensionality, and (ii) the space required to store experience would grow indefinitely as $k$ grows. Therefore, the state should contain enough features to provide a ``good summary'' of the past history and, at the same time, it should be as compact as possible to allow the learning of useful policies from a limited amount of experience. Despite that RL theory assumes the Markov property, most practical problems that may not fully satisfy this condition can be solved successfully by applying RL methods~\cite{Sutton1998}. For other cases, the MDP framework can be extended to a more general framework using partially observable MDPs (POMDPs)~\cite{Kaelbling1998}.

In the anemia management problem, the state space of the MDP was expressed as
\begin{equation*}
s=[Hb_k, \Delta Hb_k, DA_k, DA_{k-1}, DA_{k-2}, Patient_{group}]
\end{equation*}
where $Hb_k$ measures the degree of anemia in the current month $k$. $\Delta Hb_k$, defined as the difference between the current Hb level and the previous one, indicates the Hb trend. $DA_k$ is the dose of darbepoetin alfa. Due to the long-term effects of darbepoetin alfa, Hb at month $k$ is not only influenced by the last drug dose, but also by the doses during the two previous months, therefore $DA_{k-1}$ and $DA_{k-2}$ were also included in the state definition. On the other hand, given that the treatment should be tailored to the individual characteristics of each patient, information about such characteristics must be included in the state definition. The computational model employs four variables that describe each patient, namely, MCH, $Ep$, $Cp$ and $Cr$. These variables could be directly introduced as state variables, however, prior knowledge about the problem was used to reduce the state space dimensionality. MCH is a dichotomous variable that depends on the patient's gender. Therefore, it was decided to remove this variable and compute a policy for men and another for women. The remaining variables ($Ep$, $Cp$ and $Cr$) were analyzed using $k$-means clustering~\cite{Alpaydin2004}. The objective was to find groups of patients that respond to the treatment in a similar way and, then, introducing into the state definition only the information about the kind of response instead of the patient's characteristics directly. Thus, the output of the clustering algorithm, $Patient_{group}$, also formed part of the state space. The underlying assumption is that the same treatment will produce a comparable effect in patients with similar characteristics.

Regarding the action space, a set of discretized actions was used
\begin{equation*}
A=\{0,0.25,0.50,0.70,1\} \text{ $\mu$g/kg per week.}
\end{equation*}
These values correspond with the doses that can be administered to the patients. It should be noted that these weight-normalized doses are different for each patient depending on the body weight. In practice, using a discrete set of doses can be useful due to the fact that darbepoetin alfa is supplied in prefilled syringes and, if the syringe is partially injected, the rest will be wasted.


\subsection{Reward function}

The reward function defines the agent's goal. The aim of ESA treatment is to maintain Hb levels in a healthy range. According to the European Best Practice Guidelines for the Treatment of Anemia in Chronic Kidney Disease~\cite{Locatelli2004}, Hb concentration should be $>$11 g/dl for all patients, and no higher than 12 g/dl for patients with comorbidities such as cardiovascular disease, diabetes, etc. Thus, the Hb target range used in this work was [11,12] g/dl. On the other hand, abrupt changes of Hb should be avoided. Specifically, changes (increments and decrements) close to 2 g/dl per month increase the risk of cardiovascular episodes~\cite{Locatelli2004}.

In order to satisfy both goals, the reward was designed as a piecewise function that depends on the current Hb level $Hb_k$. The idea is that when $Hb_k$ is close to the target, then the policy should try to reach the target range in the next month. However, if the difference between $Hb_k$ and the target is too large, the Hb change required to reach the target in the next month can be too abrupt and it should be avoided.

\begin{figure*}[htb] 
\centering 
\subcaptionbox{\label{fig:rewardHb}}
{\includegraphics[]{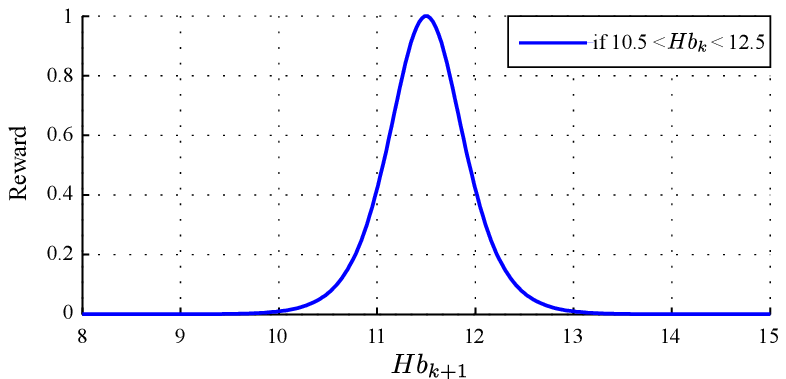}}
\qquad
\qquad
\subcaptionbox{\label{fig:rewardDeltaHb}}
{\includegraphics[]{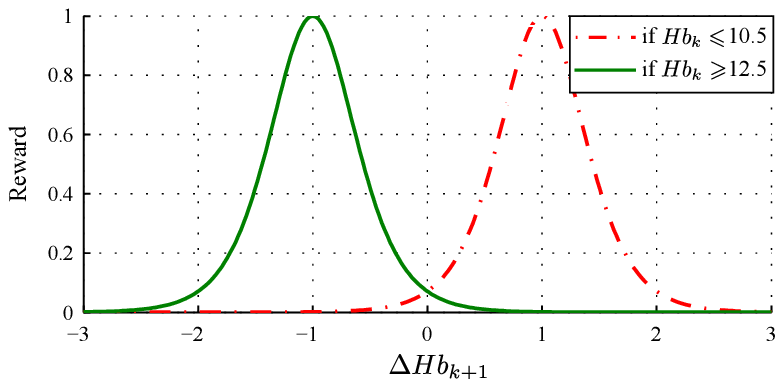}}
\subcaptionbox{\label{fig:rewardComplete}}
{\includegraphics[]{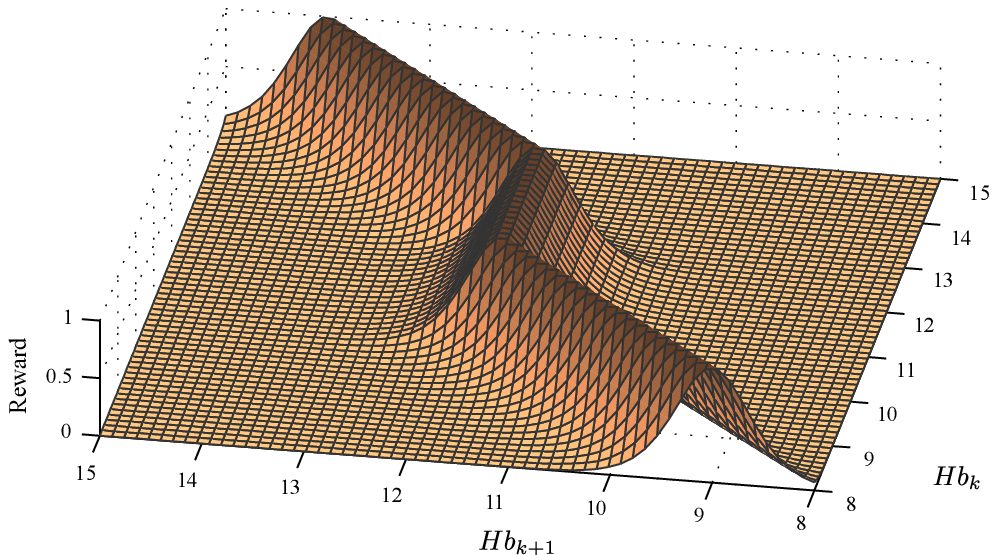}}
\caption{Reward function. Subfigures (a) and (b) represent the piecewise reward as three separate subfunctions. Function (a) is applied when $10.5<Hb_k<12.5$ and it varies depending on $Hb_{k+1}$. Functions (b) are applied when $Hb_k \leq 10.5$ (continuous green line) or $Hb_k \geq 12.5$ (dashed red line) and they vary depending on $\Delta Hb_{k+1}$. Finally, (c) represents the complete reward function versus $Hb_{k+1}$ and $Hb_{k}$.}\label{fig:reward}
\end{figure*}

The following bell-shaped reward was employed when $10.5<Hb_k<12.5$:
\begin{align}
\label{eq:rewardHb}
{\rho}_{Hb}(s_k,a_k) & = 1 - \tanh^{2}\left(\left|\frac{Hb_{k+1}-11.5}{g} \right| \cdot w\right) \\
w & = \tanh^{-1}\left(\frac{\sqrt{0.95}}{0.01}\right) \nonumber
\end{align}
where $g=0.5$ is a parameter that controls the slope of the function. This function assigns a maximum reward when $Hb_{k+1}$ is equal to 11.5 g/dl (the center of the target range), and it decreases smoothly to zero as $Hb_{k+1}$ differs from 11.5 g/dl, indicating that Hb levels near the target are preferable. Fig.~\ref{fig:rewardHb} shows the shape of ${\rho}_{Hb}$.

If $Hb_k \geq 12.5$, the goal was to achieve a decrement of 1 g/dl:
\begin{align}
\label{eq:rewardNegDelta}
{\rho}_{-\Delta Hb}(s_k,a_k) & = 1 - \tanh^{2}\left(\left|\frac{\Delta Hb_{k+1}+1}{g} \right| \cdot w\right) 
\end{align}
where $g$ and $w$ are the same as in Eq.~(\ref{eq:rewardHb}) and $\Delta Hb_{k+1}$ is the difference of Hb between the current and the next month, i.e, $Hb_{k+1} - Hb_k$. Similarly, if $Hb_k \leq 10.5$, the agent's goal was to obtain an increment of 1 g/dl. The complete piecewise reward function is given by
\begin{equation}
\label{eq:reward}
{\rho}(s_k,a_k) = \left\{
\begin{array}{lr}
       {\rho}_{Hb} &\text{if } 10.5<Hb_k<12.5\\
       {\rho}_{-\Delta Hb} &\text{if } Hb_k \geq 12.5\\
       {\rho}_{+\Delta Hb} &\text{if } Hb_k \leq 10.5\\
     \end{array}
   \right.
\end{equation}
Fig.~\ref{fig:reward} depicts each subfunction separately and the complete reward function versus $Hb_{k}$ and $Hb_{k+1}$.

%
\section{Experiments}
\label{sec:experiments}
\begin{figure*}[htb]
\begin{center} 
{\includegraphics[width=1.88\columnwidth]{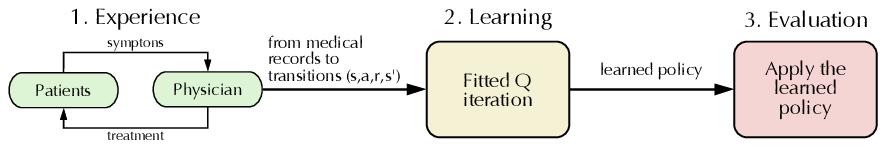}}
\caption{Experiments were carried out in three stages. First, the computational model was used to simulate a cohort of patients that are treated during a period of time, gathering the records generated by each patient in a database. Secondly, the records were processed to produce a set of transitions of the form ($s,a,r,s'$), and then FQI algorithm was applied. For comparison reasons, the learning process was also repeated using Q-learning. Third, the policies learned from the data were evaluated and compared with a standard protocol using a new cohort of simulated patients.}\label{fig:completeDiagram}
\end{center}
\end{figure*}

The purpose of the experiments was to assess the proposed method by comparing the quality of the learned policy with the popular Q-learning algorithm and with a standard protocol of dose adjustment. As illustrated in Fig.~\ref{fig:completeDiagram}, experiments were carried out in three stages. 

First, the computational model was used to generate the experience (data) necessary to apply the RL algorithm. Specifically, 5000 hemodialysis patients treated with darbepoetin alfa during 30 months were simulated. Second, the data were processed in order to obtain a set of states, actions and reward. Then, a drug administration policy was learned from data using FQI. For comparison reasons, this learning process was also carried out using Q-learning. Finally, in the third stage, a new cohort of 60 patients was simulated in accordance with three treatment strategies: the policy learned using FQI, the policy learned using Q-learning, and the treatment recommended in the protocol.

\subsection{Experience}
\label{subsec:experience}
The computational model was used to generate a data set that mimics the medical data gathered during common clinical practice. The model uses four parameters (MCH, $Ep$, $Cr$ and $Cp$) to describe the patient's individual response to anemia treatment. MCH was chosen as indicated in Eq.~(\ref{eq:MCH}), while the remaining parameters were adjusted employing data extracted from the medical records of 128 patients undergoing hemodialysis and receiving darbepoetin alfa. Data were collected between January 2008 and December 2010 in three hemodialysis centers located in Italy. Table~\ref{tab:baseline} shows the baseline characteristics of patients used to adjust the model parameters, and Table~\ref{tab:summaryParameters} shows a summary of the adjusted parameters.

\begin{table}[h]
\begin{center}
{\footnotesize
\caption{Baseline characteristics of the population used to adjust the model parameters.}\label{tab:baseline}
\begin{tabular}{p{5cm} p{2cm}}
\toprule
Characteristic & (n = 128) \tabularnewline
\midrule
Age (y) &  58.53 $\pm$ 15.15\tabularnewline
Sex (\% male) &  53.91 \tabularnewline
Height (m) &  1.62 $\pm$ 0.09 \tabularnewline
Weight (kg) &  67.97 $\pm$ 12.61 \tabularnewline
Treatment & \tabularnewline
\hspace{1.5 mm} Darbepoetin alfa ($\mu$/kg/week) & 0.29 $\pm$ 0.17 \tabularnewline
\hspace{1.5 mm} Iron IV (mg/week) &  34.74 $\pm$ 32.08 \tabularnewline 
Laboratory data & \tabularnewline
\hspace{1.5 mm} Hemoglobin (g/dl) &  11.80 $\pm$ 1.16 \tabularnewline
\hspace{1.5 mm} Ferritin (ng/ml) & 430.90 $\pm$ 178.96 \tabularnewline
\hspace{1.5 mm} Serum albumin (g/dl) &  4.09 $\pm$ 0.29 \tabularnewline
\hspace{1.5 mm} Phosphate (mg/dl) & 4.52 $\pm$ 1.03 \tabularnewline
\hspace{1.5 mm} Leukocytes (cells/mm$^3$) & 6584.1 $\pm$ 1260.9 \tabularnewline 
\bottomrule
Data are means $\pm$ SD.
\end{tabular}
}
\end{center}
\end{table}

\begin{table}[h]
\begin{center}
{\footnotesize
\caption{Summary information on the model parameters.}
\label{tab:summaryParameters}
\begin{tabular}{p{3cm} p{4cm}}
\toprule
Parameters & Value (mean $\pm$ SD) \tabularnewline
\midrule
$Ep$ &  0.3588 $\pm$ 0.0753 \tabularnewline
$Cr$ &  0.1372 $\pm$ 0.0520 \tabularnewline
$Cp$ &  0.2014 $\pm$ 0.0640 \tabularnewline
\bottomrule
\end{tabular}
}
\end{center}
\end{table}

As the proposed method was applied separately for men and women (see Section~\ref{sec:stateRepresentation}), the data set was split by gender. Due to the high level of similarity between the results obtained in both groups, the rest of the work is focused only on the men's group. After data splitting, the adjusted parameters were available for 69 male patients. In order to increase the population size, linear interpolation was applied between the parameters of each patient and a patient randomly selected among its 10 nearest neighbors. This process allows to generate artificial patients that are similar to the real ones and, at the same time, it introduces a certain degree of randomness in order to capture the variability of the characteristics among different patients. The process was repeated until the parameters corresponding to 4931 new patients were generated, making a total of 5000 patients. Then, patients' progress over time was simulated during 30 months of treatment~\footnote{Given that the prevalence of CKD is between  0.1\% and 0.2\% of the total population, the size of the simulated population is typical from a region with about 5 million inhabitants. On the other hand, patients are often treated for extended periods of time due to the chronic nature of CKD, however, 30 months was enough the generate the amount of data required to learn useful policies}.
The dose of darbepoetin alfa administered during each month was randomly selected among the discrete set $A$, which introduced a high variability of doses. If all patients had followed an specific treatment protocol, the learned policy would be strongly biased by that protocol. In the actual case, physicians' prescriptions are based on clinical protocols and in their own experience, therefore it is common to find different doses for patients with similar conditions. Nevertheless, in general, doses prescribed by physicians are closer to the optimal ones, which facilitates the learning process.

Simulations generated a total of 5000 $\times$ 30 = 150000 observations and treatments.  Due to the randomness of the treatment, in some observations the Hb level was greater than 20 g/dl. These data were removed because they represent unrealistic situations. After this restriction, the data set consisted of 138011 observations.

%

\subsection{Learning}
The data generated in the previous stage were used as input of the learning algorithms FQI and Q-learning. To this end, the sequences of observations and treatments were transformed into a set of transitions ($s$, $a$, $r$, $s^{\prime}$) following the methodology introduced in Section~\ref{sec:problemModelling}. Some of the state variables were extracted using the clustering algorithm $k$-means. The rest of this section discusses some practical issues of $k$-means, FQI and Q-learning.

\subsubsection{$k$-means}
The $k$-means algorithm employed in the clustering analysis starts with an initial estimation of $q$ centroids and then uses an iterative method to optimize the cluster quality. Typically, the number of clusters $q$ is selected empirically. In this work, the clustering analysis was repeated using values of $q$ from 3 until 10. For each value of $q$, the obtained clusters were analyzed using silhouette plots~\cite{Rousseeuw1987}; finally, $q=5$ was selected as the most suitable because it maximizes the within-cluster compactness and the separability among clusters.

\subsubsection{Fitted Q iteration}
FQI is based on an iterative procedure and the designer must decide when to stop the iterations. The algorithm starts with an arbitrary approximation of the optimal Q-function that is improved at each iteration. Thus, the number of iterations should be enough to allow convergence to the optimal Q-function. 
Convergence can be measured in terms of distance between consecutive approximations of the Q-function, defined as
\begin{equation}
\label{eq:convergence}
\text{dist}(\hat{Q}_n,\hat{Q}_{n-1})= \frac{\sum_{(s,a) \in S^i \times A}\left(\hat{Q}_n(s,a) - \hat{Q}_{n-1}(s,a)\right)^2}{\#(S^i \times A)}
\end{equation}
where $S^i \times A$ are the state-actions pairs contained in the set of transitions~\cite{Ernst2005}. 

\begin{figure}
\begin{center} 
{\includegraphics[]{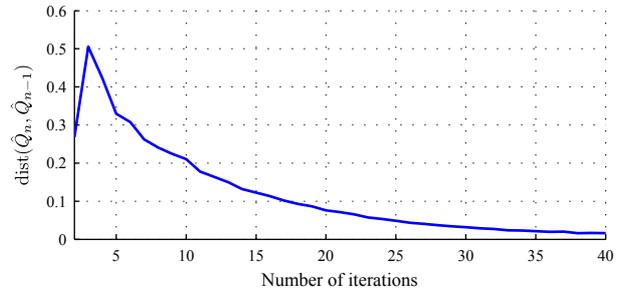}}
\caption{Fitted Q iteration convergence measured as the distance between $\hat{Q}_n$ and $\hat{Q}_{n-1}$ versus number of iterations.}\label{fig:convergence}
\end{center}
\end{figure}

Fig.~\ref{fig:convergence} shows the convergence of FQI versus the number of iterations, where it can be observed that the approximated Q-function remains almost stable after 35 iterations. In view of this, the number of iterations was fixed to 40. On the other hand, the discount factor $\gamma$ was empirically set to 0.9, this value was enough to incorporate the long-term effects of each dose in the optimization process.

Extra-Tree algorithm parameters were chosen following the procedure described in~\cite{Ernst2005}. The number of trees in the ensemble was set to $M=50$, the parameter $K$ was set equal to the dimension of the input space, i.e, $K = 6$, and the minimal leaf size was selected among the values $l_{min}=[5,10,50,100]$ using cross-validation in each iteration of FQI. Experiments showed that these default values are a good choice~\cite{Geurts2005}. 

\subsubsection{Q-learning}
Q-learning is also based on an iterative procedure that estimates the optimal Q-function. However, as each transition $(s,a,r,s')$ is only used one time, when the number of transitions is limited the algorithm simply iterates over the complete set of transitions. Thus, it is not necessary to implement a strategy to stop the iterative procedure.

The Gaussian RBF network with fixed bases employed to approximate the Q-function requires the definition of the number of Gaussian functions, their centers and standard deviations.
This process typically requires trial and error experimentation with various configurations. The number of functions should be enough to provide a smooth interpolation over the entire state-action space. If the Gaussians are too peaked, it will be necessary to employ a very large number of functions to cover the entire space. On the other hand, if they are too flat, the RBF network will not be flexible enough to approximate abrupt changes of the Q-function. Thus, it is necessary to reach a compromise between locality and smoothness.

\begin{figure}
\begin{center} 
{\includegraphics[]{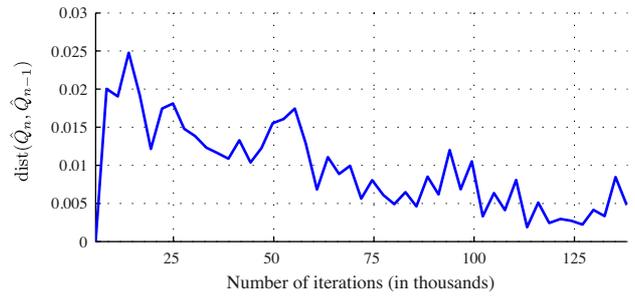}}
\caption{Q-learning convergence measured as the distance between $\hat{Q}_n$ and $\hat{Q}_{n-1}$ versus number of iterations.}\label{fig:convergenceQlearning}
\end{center}
\end{figure}

After experimenting with several configurations, the selected RBF network architecture employed 4096 Gaussian functions whose centers were distributed in a regular grid over the state space. The inputs were normalized to the range $[-1,1]$ and all the standard deviations were set to $\sigma=1.1$. The discount factor was fixed as in the FQI algorithm ($\gamma = 0.9$). Additionally, Q-learning introduces a new parameter that should be tuned: the learning rate $\alpha$. A large value of $\alpha$ usually results in faster convergence, but when it exceeds a certain critical value, the algorithm becomes unstable. Fig.~\ref{fig:convergenceQlearning} shows the convergence of Q-learning (measured using Eq.~(\ref{eq:convergence})) when $\alpha = 0.2$. The figure suggests that Q-learning has not converged after 138011 iterations. 

It is worth noting that each iteration of Q-learning (Fig.~\ref{fig:convergenceQlearning}) employs only a transition of the data set; whereas in FQI (Fig.~\ref{fig:convergence}) each iteration is based on the complete data set. Hence, despite the number of iterations shown in both figures, the convergence process of FQI is much larger than in the case of Q-learning.

\subsection{Evaluation}
The policy learned with the proposed methodology based in FQI was evaluated in a cohort of 60 new patients using the computational model. Similar to Section~\ref{subsec:experience}, the parameters corresponding to these patients were generated by linear interpolation. The evolution of each patient was simulated during 30 months of treatment using the drug doses indicated by the RL policy. For comparison purposes, the same cohort of patients was simulated according to the policy learned with Q-learning and the treatment recommended by a standard protocol. The protocol was extracted from the European Medicines Agency~\cite{EMA2013} and it describes the dosage regimen for Aranesp$^{\text{TM}}$ (a commercial form of darbepoetin alfa) as follows:

\begin{itemize}
\item The initial dose for patients on dialysis should be 0.45 $\mu$g/kg body weight administered weekly.
\item If the increase in Hb is inadequate (less than 1 g/dl in four weeks) the dose should be increased by approximately 25\%. 
\item The dose must not be increased more frequently than once every 4 weeks. 
\item If the Hb rises more than 2 g/dl in a period of 4 weeks, the dose should be reduced by 25\%.
\item If the Hb level is greater than 12 g/dl, the dose should be reduced by 25\%. After this reduction, if the Hb level continues rising, the treatment should be temporarily interrupted until the level starts to decline, moment in which the treatment should be resumed with a dose approximately 25\% lower than the previous dose.
\end{itemize}

When the computational model that simulates the patients is initialized, the level of Hb is determined by the initial conditions. These conditions are equal for all the patients (see~\ref{sec:appendix}), i.e., at month zero all the patients have the same Hb level. In order to have a cohort of patients with a heterogeneous initial state, the 60 patients were treated during four months with a random treatment before starting the evaluation process.  Thus, the protocol recommendation for the initial dose was never applied because the patients had already received some previous doses of ESA when beginning the evaluation.



\section{Results and discussion}
\label{sec:results}
This section assesses the proposed methodology by comparing the policy learned using FQI ($\pi_{FQI}$) with the policy learned using Q-learning ($\pi_{Q\text{-}learning}$) and extracted from the protocol ($\pi_{protocol}$). All results shown correspond with the cohort of 60 simulated patients used for validation purposes and 30 months of treatment.

\begin{figure}[]	
\begin{center} 
{\includegraphics[]{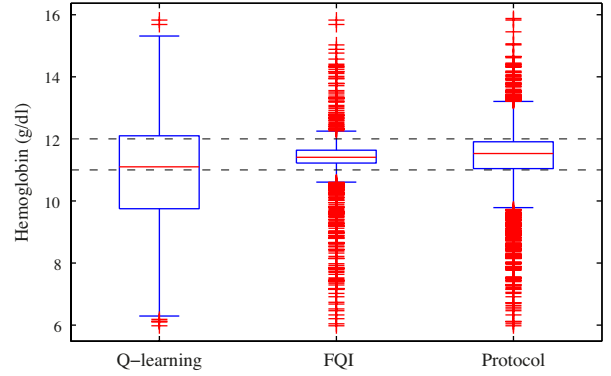}}
\caption{Box plot representation of the Hb levels corresponding to 60 simulated patients during 30 months of treatment. Patients were treated according to $\pi_{Q\text{-}learning}$, $\pi_{FQI}$ and $\pi_{protocol}$. The two horizontal lines (dashed black) indicate the Hb target range.}\label{fig:boxplot}
\end{center}
\end{figure}

As a first approach to compare the behavior of the three policies, the Hb levels obtained with each policy were represented in a box plot. This kind of representation graphically depicts a set of data values through their quartiles, allowing to visually estimate the degree of dispersion and skewness. Fig.~\ref{fig:boxplot} shows the box plot corresponding to $\pi_{Q\text{-}learning}$, $\pi_{FQI}$ and $\pi_{protocol}$. In the three cases the median falls within the desired range of Hb (indicated by two horizontal dashed lines at 11 and 12 g/dl) and the Hb levels are approximately symmetrically distributed. The main difference observed among the three policies is the large dispersion of $\pi_{Q\text{-}learning}$, which means that many of the patients treated with this policy have Hb levels away from the target range. In the other two policies there are also Hb levels substantially different from the target range; however, these values can be considered outliers (shown as red crosses) from the complete set of observations and are likely due to the initial states of the patients.

\begin{figure*}[t] 
\centering 
\subcaptionbox{\label{fig:evalRL}}
{\includegraphics[]{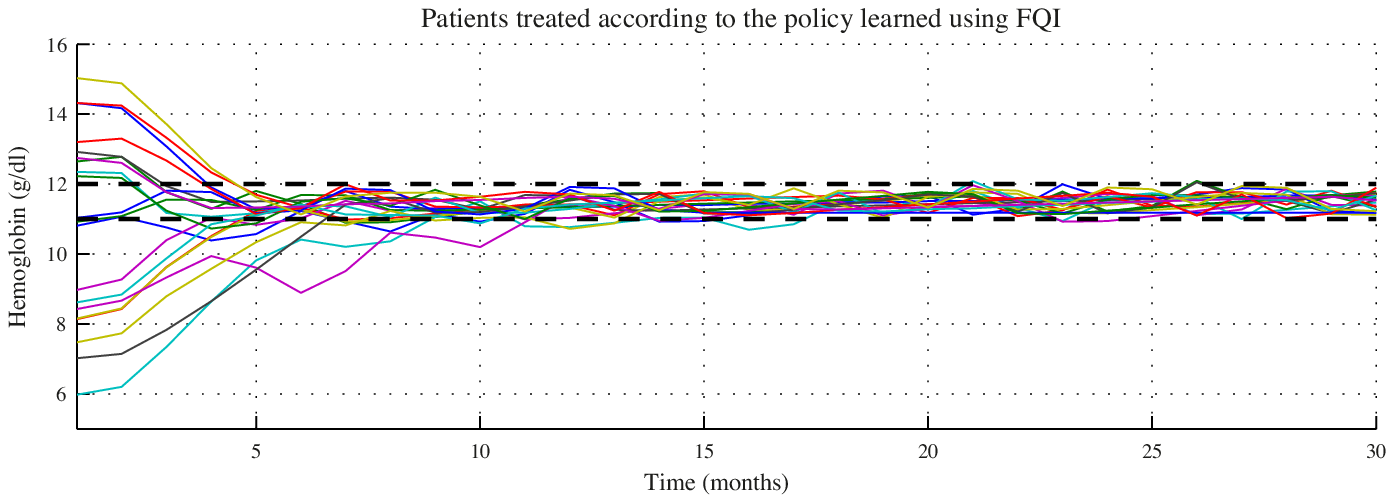}}
\qquad
\qquad
\subcaptionbox{\label{fig:evalAgmen}}
{\includegraphics[]{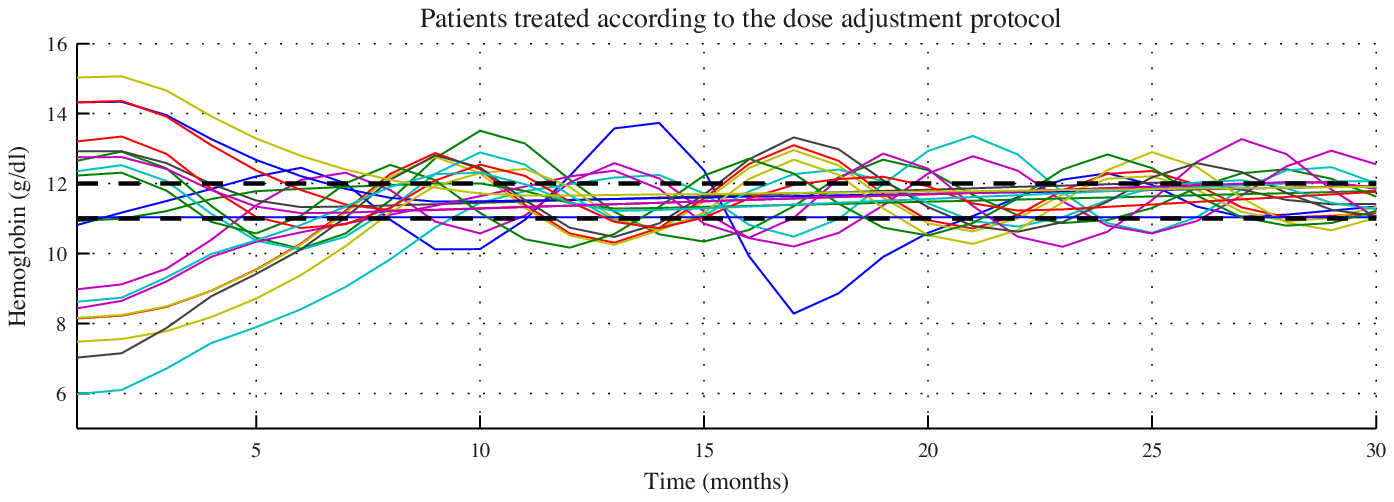}}
\caption{Hb level evolution during 30 months of treatment for 19 simulated patients randomly chosen from the test group. In (a), the patients were treated according to the policy learned with the proposed method, $\pi_{FQI}$. While in (b) they were treated according to the policy extracted from the protocol, $\pi_{protocol}$. The two horizontal lines (dashed black) indicate the Hb target range.}\label{fig:evaluation}
\end{figure*}

The low performance of Q-learning is probably a consequence of the limited amount of transitions available to estimate the policy. In principle, both RL algorithms (Q-learning and FQI) should be able to find similar policies in a given problem. In fact, other authors found that, given enough data, the policy learned by Q-learning was superior than the protocol~\cite{Gaweda2005, Martin2009}. Nonetheless, a key advantage of FQI is that it makes a more efficient use of the data, which can be crucial in problems where obtaining data is a non-trivial task. A second factor that may be related with the superiority of FQI is its flexibility to approximate the Q-functions. In Q-learning it is necessary to define an a priori approximator structure that remains fixed over the learning process. On the contrary, in each iteration FQI selects the approximator structure that provides the best approximation of the current Q-function. 


Since FQI outperforms Q-learning considerably, the rest of this section is focused only on analyzing the differences between the policy obtained using FQI and the protocol of dose adjustment.

A limitation of the box plot representation is that the temporal information is lost. In order to compare the performance of the two policies along time, a better approach consists in representing the Hb levels during the treatment. Fig.~\ref{fig:evaluation} shows this information for $\pi_{FQI}$ and $\pi_{protocol}$. For the sake of simplicity, Fig.~\ref{fig:evaluation} represents the Hb of only a subset of 19 patients randomly selected. The figure also includes two horizontal lines (dashed black) at 11 and 12 g/dl that indicate the target range. During the first months of treatment, Hb was shifted towards the target range with either of the two policies. In this sense, $\pi_{FQI}$ was more effective than $\pi_{protocol}$ because, in general, it required less time to reach the target. After seven months of treatment, several signs of Hb cycling appeared in some of the patients treated with $\pi_{protocol}$. On the contrary, $\pi_{FQI}$ was able to prevent or drastically reduce Hb cycling. As expected, the extreme values of Hb previously shown as outliers in the box plot are mainly concentrated in the first months of the treatment.

\begin{figure}[t]
\begin{center} 
{\includegraphics[]{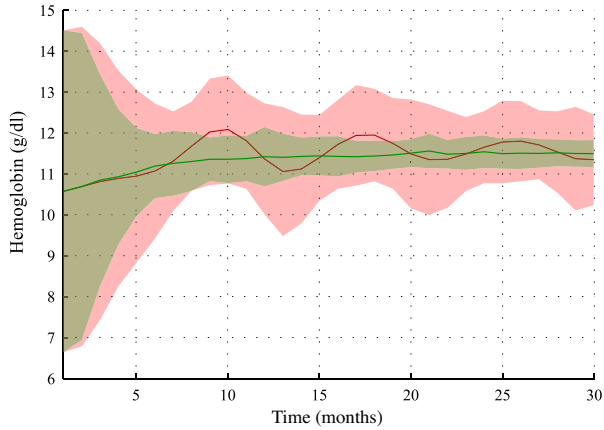}}
\caption{Mean Hb level (solid lines) and standard deviation (shaded areas) over time for 60 patients simulated according $\pi_{FQI}$ (in green) and $\pi_{protocol}$ (in red).}\label{fig:meanHb}
\end{center}
\end{figure}

Fig.~\ref{fig:meanHb} also shows the variation of Hb over time, but in this case in terms of the mean (solid lines) and standard deviation (shadow areas) for the complete group of patients. Again, it can be observed that $\pi_{FQI}$ stabilizes the Hb levels within the target range whereas the protocol produces oscillations. The large standard deviation corresponding to $\pi_{protocol}$ indicates that there are patients with dangerous Hb levels in all months. Although the Hb cycling produced by $\pi_{protocol}$ diminishes slightly over time, after 30 months it is still noticeable. Hb variability in patients treated with $\pi_{FQI}$ was significantly lower (p $<$ 0.0001, one-tailed paired F-test) than those treated with $\pi_{protocol}$.

\begin{figure}[t]
\begin{center} 
{\includegraphics[]{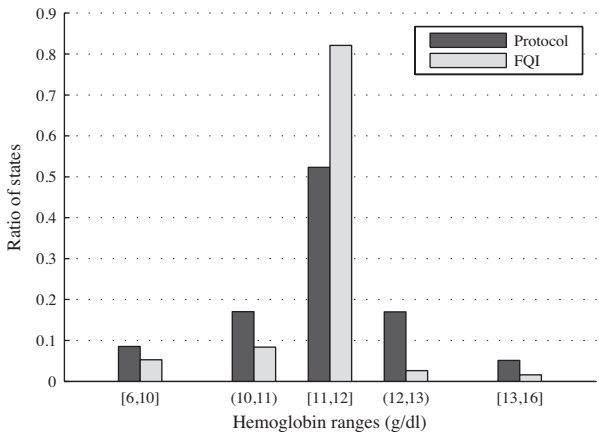}}
\caption{Comparison of the observed monthly Hb for the patients simulated following $\pi_{FQI}$ and $\pi_{protocol}$ during 30 months. Ideally, all Hb observations should be in the range $[11,12]$ g/dl.}\label{fig:histogram}
\end{center}
\end{figure}

A second metric employed to compare both policies was the number of months or observations in which the patients presented an adequate Hb level. Fig.~\ref{fig:histogram} presents this information by means of a bar chart. Five ranges of Hb grouped into three categories were defined: the category \emph{suitable} includes the target range [11,12] g/dl; \emph{unsuitable} consists of the two ranges contiguous to the target range, that is, (10,11) and (12,13) g/dl; and \emph{dangerous} contains the rest of Hb levels. Fig.~\ref{fig:histogram} shows two bars corresponding with $\pi_{FQI}$ (light gray) and $\pi_{protocol}$ (dark gray) in each one of the five ranges. Ideally, all patient observations should be within the category \emph{suitable}. When the patients were treated with $\pi_{FQI}$, their Hb level was within the target range in 82.1\% of the observations, which represents an important improvement compared with the 54.5 \% achieved by $\pi_{protocol}$. The rest of patient observations were mainly located in the category \emph{unsuitable}, specifically, 11.02\% for $\pi_{FQI}$ and 34.1\% for $\pi_{protocol}$. This percentage indicates that $\pi_{protocol}$ also achieved a reasonable outcome. Finally, a minor percentage of patient observations was in the category \emph{dangerous}; however, Figs.~\ref{fig:evaluation} and \ref{fig:meanHb} shows that these observations corresponds mainly with the initial months of treatment, when the drug administered has not yet produced any effect.

In addition to stabilizing the Hb level inside the target range, the treatment must avoid abrupt Hb changes. The difference between $\pi_{FQI}$ and $\pi_{protocol}$ concerning this aspect was negligible. Both policies were really effective in avoiding Hb changes greater than 2 g/dl per month, more than 99.5\% of state transitions met this condition. 

Due to the high costs of darbepoetin alfa, another point to take into account when comparing both policies is the quantity of drug used. There was a highly significant difference ($\text{p}<0.0001$, one-tailed paired t-test) of 5.13\% between the mean dose recommended by $\pi_{protocol}$ (0.39 $\mu$g/kg/week) and $\pi_{FQI}$ (0.37 $\mu$g/kg/week). Thus, the treatment recommended by the RL policy not only produced better outcomes but also generated lower costs.

Finally, a particular case was studied in detail to obtain more insight into the behavior of each policy. Table~\ref{tab:results} shows how the Hb level of a patient and the dose of drug administered varies along 15 months of treatment. The goal of this table was to compare the treatment recommended by both policies in a case where $\pi_{protocol}$ caused Hb cycling. The months where Hb level was within the target range are highlighted in bold. It can be observed that, in the fifth month, $\pi_{protocol}$ started to increase the darbepoetin dose because $Hb < 11$ g/dl, and it continued increasing the dose in months 6 and 7 because the Hb level had not yet reached the target range. Then, after reaching the target (month 8), the Hb level continued rising despite drug dose was decreased (months 9 and 10). This phenomenon is due to the delay between drug administration and its effects in Hb. On the other hand, the doses administered by $\pi_{FQI}$ suggest that the long-term effects are taken into account. For example, in months 7 and 8 the dose is maintained constant despite the low Hb level. In consequence, the Hb level in the next months increases and is stabilized within the target.

\begin{table}[h]
\begin{center}
{\footnotesize
\caption{Comparison of the treatment and clinical evolution of a patient treated according to $\pi_{FQI}$ and $\pi_{protocol}$. The Hb levels within the target range are highlighted in bold.}\label{tab:results}
\begin{tabular}{p{1cm}  p{1cm} p{1.4cm} p{1cm} p{1.4cm}}
\toprule
Month & \multicolumn{2}{c}{FQI} & \multicolumn{2}{c}{Dose adjustment protocol} \tabularnewline
\cmidrule(r){2-3} \cmidrule(r){4-5}
 & Hb (g/dl) & Drug dose ($\mu$g/kg/week) & Hb (g/dl) & Drug dose ($\mu$g/kg/week)\tabularnewline
\midrule
1 & 12.65 & 0.50 & 12.65 & 0.75\tabularnewline
2 & 12.78 & 0.75 & 12.91 & 0.56\tabularnewline
3 & \textbf{11.77} & 0.75 & 12.43 & 0.42\tabularnewline
4 & \textbf{11.28} & 0.50 & \textbf{11.36} & 0.42\tabularnewline
5 & \textbf{11.80} & 0.50 & 10.44 & 0.52\tabularnewline
6 & \textbf{11.43} & 0.50 & 10.13 & 0.65\tabularnewline
7 & 10.89 & 0.50 & 10.57 & 0.82\tabularnewline
8 & 10.91 & 0.50 & \textbf{11.52} & 0.82\tabularnewline
9 & \textbf{11.06} & 0.50 & 12.72 & 0.61\tabularnewline
10 & \textbf{11.22} & 0.50 & 13.50 & 0.46\tabularnewline
11 & \textbf{11.39} & 0.50 & 13.14 & 0.34\tabularnewline
12 & \textbf{11.56} & 0.50 & 12.11 & 0.34\tabularnewline
13 & \textbf{11.73} & 0.25 & \textbf{11.24} & 0.34\tabularnewline
14 & \textbf{11.74} & 0.50 & 10.90 & 0.43\tabularnewline
15 & \textbf{11.19} & 0.25 & \textbf{11.00} & 0.54\tabularnewline
\bottomrule
\end{tabular}
}
\end{center}
\end{table}

The experiments reported in this paper were based on a computational model that simulates the patients' response to the treatment with darbepoetin alfa. Although the proposed methodology is valid for the case of actual patients, certain aspects should be modified. The main differences between simulated and real patients are discussed below.

The inclusion of the model parameters (MCH, $Ep$, $Cp$, $Cr$) in the definition of the state space is likely the most important difference between applying the proposed system to simulated and real patients. These parameters, which are related to the patients' individual characteristics, are used to find groups of patients that respond to the treatment in a similar way. In practice, the parameters cannot be directly measured. One option would be to adjust them as in the simulated case; however, a more effective solution is to employ the variables commonly measured in the monthly reviews which provide the same information. For example, the level of C-reactive protein, serum albumin and leukocytes can be used as an indicator of inflammation level. Thus, the clustering analysis should be performed using those variables as inputs instead of the model parameters.

Some elements of the MDP, such as the reward function and the discount factor, should be carefully chosen to provide the desired outcomes. When using a computational model, trial-and-error procedures can be used to adjust them. On the contrary, this approach is not viable in real domains. Thus, despite the valuable experience gained with simulated patients, the expertise and advice of physicians are still necessary to design the MPD in the real case.

Finally, a third issue that should be noticed is the two assumptions made by the model: stable level of inflammation and constant iron availability. As previously mentioned, these assumptions may are not met in the real case. The only requirement to overcome both assumptions is that the clinical data employed to learn the policy must contain enough examples of patients  in which assumptions are violated, i.e, with variable levels of inflammation and iron availability.


\section{Conclusions and future work}
\label{sec:conclusions}

This work has proposed a methodology based on RL to optimize anemia treatment in hemodialysis patients. After formulating the problem using the MDP framework, RL was able to learn automatically near-optimal treatments from clinical data. Contrary to other techniques to solve MPDs, RL does not require a complete knowledge of the system dynamics, a feature that can be crucial in medical problems. More specifically, the methodology uses the algorithm FQI, which stands out for its ability to make an efficient use of data. FQI was combined with a function approximator based on regression trees in order to deal with the continuous state space and to generalize the learned policy to the cases not covered by the data set. The state variables of the MDP were extracted partly from the laboratory tests and partly from a clustering analysis of the patient's main attributes.


The proposed methodology was evaluated through a computational model that describes the effect of darbepoetin alfa on the concentration of Hb. In addition to FQI, the experiments were also performed using the well-known Q-learning algorithm. A standard protocol of dose adjustment was used for baseline comparison. The quality of the policy obtained with Q-learning, $\pi_{Q\text{-}learning}$ was considerably inferior in comparison to the other two policies ($\pi_{FQI}$ and $\pi_{protocol}$). When comparing $\pi_{FQI}$ and $\pi_{protocol}$, the policy obtained with FQI increased by 27.6\% the proportion of patients with an adequate level of Hb and, at the same time, it reduced the amount of drug used by 5.13\%.


The simulation results suggest that the FQI policy can deal with the long-term effects of darbepoetin alfa and the high variability in the patient's response. These features, which are absent in standard dosing protocols, had been suggested as a major cause of Hb cycling. As a result, the proposed methodology was more effective than the standard-use tested protocol in maintaining stable Hb levels and preventing Hb cycling. On the other hand, the drug is prescribed in a more efficient way since the treatment achieves better outcomes with less amount of darbepoetin alfa.


The computational model used in the experiments has several limitations owing to the assumptions on which it is grounded and, therefore, does not represent all possible patients. Nevertheless, it reproduces some important difficulties present in actual cases that may cause Hb cycling. Thus, although prospective validation is required, experiments have shown the potential benefits of RL in anemia treatment.

The positive results obtained in this work using simulated patients have motivated further research in applying the proposed methodology to actual patients. Currently, a tool for clinical decision support based on FQI is being validated through a clinical evaluation in five hemodialysis centers from three European countries.


Although this work has been focused in renal anemia, the methodology can be extended to other types of anemia. For example, oncology patients who also receive darbepoetin alfa treatment, or even to other complex problems of drug administration such as warfarin therapy to prevent venous thromboembolism. Another interesting line of future research is to include other optimization aspects in the reward function, in addition to those related to the patients' health, like the cost of the treatment. 

\section*{Acknowledgements}
The authors thank the reviewers from Artificial Intelligence in Medicine Journal, whose insightful and helpful comments were of great value for the preparation of the final manuscript.

\appendix
\section{Erythropoiesis model}
\label{sec:appendix}

Many mathematical models have been developed to simulate the process of erythropoiesis in different physiological scenarios~\cite{Ramakrishnan2004, Woo2006, Krzyzanski2005}. The most common approach is the use of age-structured models~\cite{Belair1995, Banks2004, Ackleh2006}. The model presented in this section is focused in patients undergoing hemodialysis who are treated with intravenous darbepoetin alfa. In addition, it incorporates the effects of iron availability and level of inflammation, although both are considered constant. Similar to~\cite{Krzyzanski1999}, the stimulating action of the drug is described through a multi-compartment model. Three different cell classes or compartments are considered:
\begin{itemize}
\item $P$ is the bone marrow concentration of progenitor cells (BFU-E and CFU-E) and precursor cells (proerythroblast and basophilic erythroblast) that depend on EPO for their survival and differentiation to the next compartment.
\item $M$ comprises the remaining erythroblasts (polychromatic and orthochromatophilic) and marrow reticulocytes which are iron-responsive. 
As it is assumed that all patients have sufficient iron available, this compartment only serves as delay on the differentiation until the next compartment. 
\item $R$ is the plasma concentration of red blood cells.
\end{itemize}

The equations governing the evolution of the number of cells in each population are simply balance equations~\cite{Krzyzanski1999}. Each compartment of cells is fed by an entering flow of fresh cells and is emptied by the outgoing flow of differentiated or apoptotic cells.

The fundamental assumption is that every cell in each population lives for the same period of time, which is constant and denoted by ${\bf T}_P$ for cells $P$ and  ${\bf T}_R$ for cells $R$~\cite{Krzyzanski1999}. This assumption determines the cell elimination rate since the number of cells that are lost at time $t$ must be equal to the number of cells that are born at the same time delayed by the appropriate lifespan.
The loss process is modeled by means of weighted averages of the previous day incoming rates in order to take into consideration that cells actually have different exposition times to the drug. Such exposition time varies according to their internal maturity level at the time of the administration.

The flow entering the $P$ compartment depends on the progenitors response to the stimulatory effect of erythropoietin. According to~\cite{Krzyzanski1999}, this response can be described by the Hill function
$
H(E_{\rm tot}):= E_{\rm tot} / (E_{50} + E_{\rm tot}),
$
where  $E_{\rm tot}$, which is defined as in Eqs.~(\ref{epotot}) and (\ref{epoexo}),  represents the total plasma concentration of EPO, and $E_{50}=100/V_d$. The sensitivity $E_{50}$ is the half-maximal effective concentration, a parameter commonly used as a measure of drug's potency.

The outgoing flow of the $P$ compartment is also the feeding flow to the $R$ compartment delayed by ${\bf T}_M$, as it is assumed that, up to a proportionality factor (describing a proliferation activity of $M$ cells), as many $M$ cells as $P$ cells are produced.

At the last stage of their lifespan, RBCs become senescent and are cleared from the blood. The rate at which RBCs are cleared is a weighted average of the previous incoming rates. Such a average is a convex combination with coefficients given by a Gaussian distribution with mean equal to the RBC mean lifespan and variance $30$. The Gaussian distribution indicates that the drug effects are stronger on ${\bf T}_R$-day old cells. 

The evolution laws of the compartment $P$ and $R$ are the following: 
\begin{equation*}
\left\{\begin{array}{l}
P'(t) = C_p \dfrac{E_{\rm tot}(t)}{E_{50} +E_{\rm tot}(t)} P(t)
\vspace{.2truecm}\\
  \qquad        - C_p  \dfrac{1}{{\bf T}_P} \displaystyle\sum_{T_j=1}^{{\bf T}_P} 
           \dfrac{E_{\rm tot}(t-T_j)}{E_{50} +E_{\rm tot}(t-T_j)} P(t-T_j)
\vspace{.2truecm}\\
R'(t) = C_r \dfrac{E_{\rm tot}(t- ({\bf T}_P+{\bf T}_M))}{E_{50} +E_{\rm tot}(t-({\bf T}_P+{\bf T}_M))} 
P(t-({\bf T}_P+{\bf T}_M))
\vspace{.2truecm}\\
\qquad           -   \dfrac{C_r}{S_{P,M,R}} 
 \displaystyle\sum_{T_j={\bf T}_P+{\bf T}_M+1}^{{\bf T}_P+{\bf T}_M +{\bf T}_R} 
           g_{T_j} \dfrac{E_{\rm tot}(t-T_j)}{E_{50} +E_{\rm tot}(t-T_j)} P(t-T_j)
\end{array}\right.
\end{equation*}
where $g_{T_j}:=G(T_j -({\bf T}_P+{\bf T}_M)))$, $G(T)$ is 
the Gaussian distribution with mean ${\bf T}_{R}$ and variance $30$
           evaluated at time $T$, and 
           $S_{P,M,R}$ is a normalization factor given by
\[S_{P,I,R}:=\displaystyle\sum_{T_j={\bf T}_P+{\bf T}_M+1}^{{\bf T}_P+{\bf T}_M +{\bf T}_R} 
           G\Big(T_j- ({\bf T}_P+{\bf T}_M)\Big) 
           =\sum_{T_j=1}^{{\bf T}_R} 
           G\big(T_j\big).\]
           
In the present study,  $({\bf T}_P, {\bf T}_M,{\bf T}_R)= (9,4,70)$. At the time $t_0$ of the first administration, $P_0=1$ and $R_0=1$ (i.e., there are $10^{11}$ cells maturing from class $P$ to $R$). The remaining parameters, $Ep$, $Cp$ and $Cr$, are estimated for each patient. The solution of the model, given in Eq.~(\ref{hgmodsol}), is fitted in the least square sense using the first six Hb levels of each patient measured through laboratory tests. Then, the model can be used to simulate the response of that patient to the treatment.

\begin{figure}[tb]
\begin{center} 
{\includegraphics[]{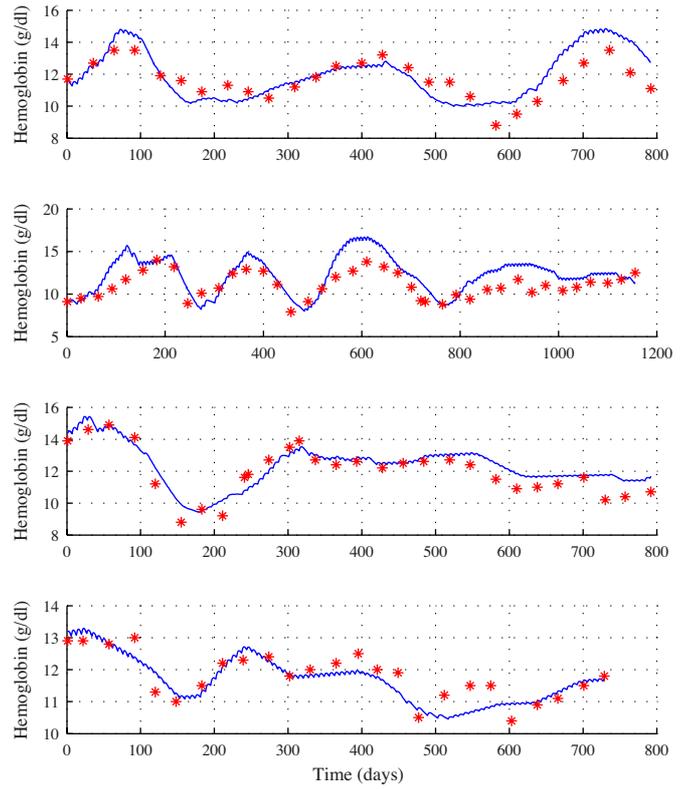}}
\caption{Hb level simulated using the model (blue line) and measured by laboratory tests (red asterisks) corresponding to four patients during approximately 26 months of treatment. In the four cases, model's assumptions are approximately met.}\label{fig:modelResponse}
\end{center}
\end{figure}

The basic assumptions of the model are that each patient maintains a stable level of inflammation and that iron availability for erythropoiesis is constant. However, dialysis patients often have multiple comorbidities (such as hypertension, diabetes or cardiovascular disease) that may produce fluctuating levels of inflammation and reduce the availability of iron, which limits the usefulness of the model. 

On the other hand, when the assumptions are met, the model is able to reproduce the variability among patients in drug response and the long-term effect produced by the drug. Fig.~\ref{fig:modelResponse} shows four cases in which the assumptions are reasonably well met. As it can be observed, in general, the model (continuous blue line) is able to simulate the Hb level measured by laboratory tests (res asterisks).




\bibliographystyle{model3-num-names}
\bibliography{bibliography}







\end{document}